\documentclass{article} % For LaTeX2e
\usepackage{iclr2024_conference,times}

\usepackage{amsmath,amsfonts,bm}

\def\eqref#1{equation~\ref{#1}}
\def\1{\bm{1}}

\DeclareMathAlphabet{\mathsfit}{\encodingdefault}{\sfdefault}{m}{sl}
\SetMathAlphabet{\mathsfit}{bold}{\encodingdefault}{\sfdefault}{bx}{n}

\DeclareMathOperator*{\argmax}{arg\,max}

\usepackage{hyperref}
\usepackage{url}
\usepackage{times}
\usepackage{latexsym}
\usepackage{float}
\usepackage{amsthm}
\usepackage{tcolorbox}
\usepackage{microtype}
\usepackage{amsmath, amssymb}
\usepackage{booktabs}
\usepackage{wrapfig}

\usepackage{graphicx}
\usepackage{multirow}

\usepackage{bm}
\usepackage{algorithm}
\usepackage{pifont}
\usepackage{subfigure}
\usepackage{xspace}

\title{\methodname: Verify and Reinforce LLMs Step-by-step without Human Annotations}

\author{Peiyi Wang$^{1}$$\footnotemark[2]$ \quad Lei Li$^3$ \quad Zhihong Shao$^4$ \quad R.X. Xu$^2$ \quad Damai Dai$^1$ \quad Yifei Li$^5$\\
\textbf{Deli Chen$^2$  \quad Y. Wu$^2$ \quad Zhifang Sui$^1$} \\ 
$^1$National Key Laboratory for Multimedia Information Processing, Peking University\\
$^2$DeepSeek-AI \quad $^3$The University of Hong Kong \\
$^4$Tsinghua University  \quad $^5$The Ohio State University \\
 \texttt{\{wangpeiyi9979, nlp.lilei\}@gmail.com}  \\
 \texttt{{li.14042}@osu.edu} \quad \texttt{szf@pku.edu.cn}
}

\newcommand{\methodname}{\textsc{Math-Shepherd}\xspace}

\begin{document}
\renewcommand{\thefootnote}{\fnsymbol{footnote}}
\footnotetext[2]{Contribution during internship at DeepSeek-AI.}
\renewcommand{\thefootnote}{\arabic{footnote}}

\maketitle

\vspace{-8.5mm}
\begin{center}
         \fontsize{9.5pt}{\baselineskip}\selectfont {\includegraphics[height=7mm]{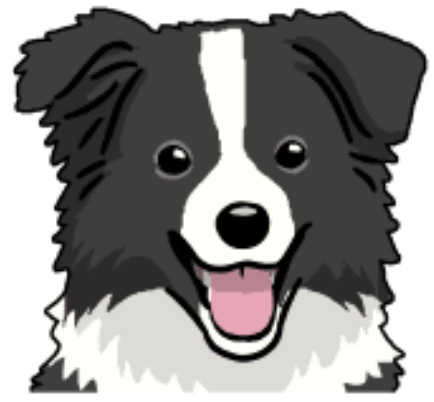}} ~
         {Project Page:}~\tt\href{https://achieved-bellflower-4d6.notion.site/Math-Shepherd-Verify-and-Reinforce-LLMs-Step-by-step-without-Human-Annotations-41b6e73c860840e08697d347f8889bac?pvs=4}{\methodname}
         \vskip 0.25in
\end{center}

\begin{abstract}
In this paper, we present an innovative process-oriented math process reward model called \textbf{\methodname}, which assigns a reward score to each step of math problem solutions. 
The training of \methodname is achieved using automatically constructed process-wise supervision data, breaking the bottleneck of heavy reliance on manual annotation in existing work. 
We explore the effectiveness of \methodname in two scenarios: 
1) \textit{Verification}:  \methodname is utilized for reranking multiple outputs generated by Large Language Models (LLMs);
2) \textit{Reinforcement Learning}: \methodname is employed to reinforce LLMs with step-by-step Proximal Policy Optimization (PPO).
With \methodname, a series of open-source LLMs demonstrates exceptional performance. 
For instance, the step-by-step PPO with \methodname significantly improves the accuracy of Mistral-7B (77.9\%$\to$84.1\% on GSM8K and 28.6\%$\to$33.0\% on MATH).
The accuracy can be further enhanced to 89.1\% and 43.5\% on GSM8K and MATH with the verification of \methodname, respectively.
We believe that automatic process supervision holds significant potential for the future evolution of LLMs.

\end{abstract}
% add citation 
% \begin{quote}
%   \textit{``Education is not the learning of facts, but the training of minds to think.'' --- Albert Einstein}
% \end{quote}

\begin{figure}[h]
\centering
\subfigure{
\includegraphics[width=0.435\textwidth]{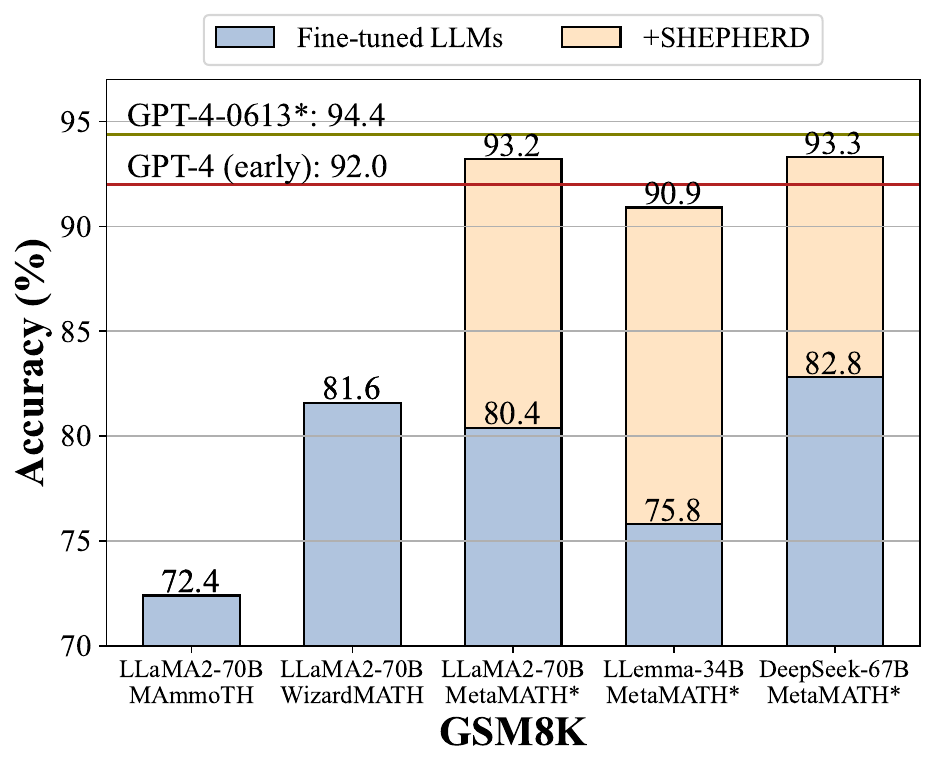}
}
\subfigure{
\includegraphics[width=0.435\textwidth]{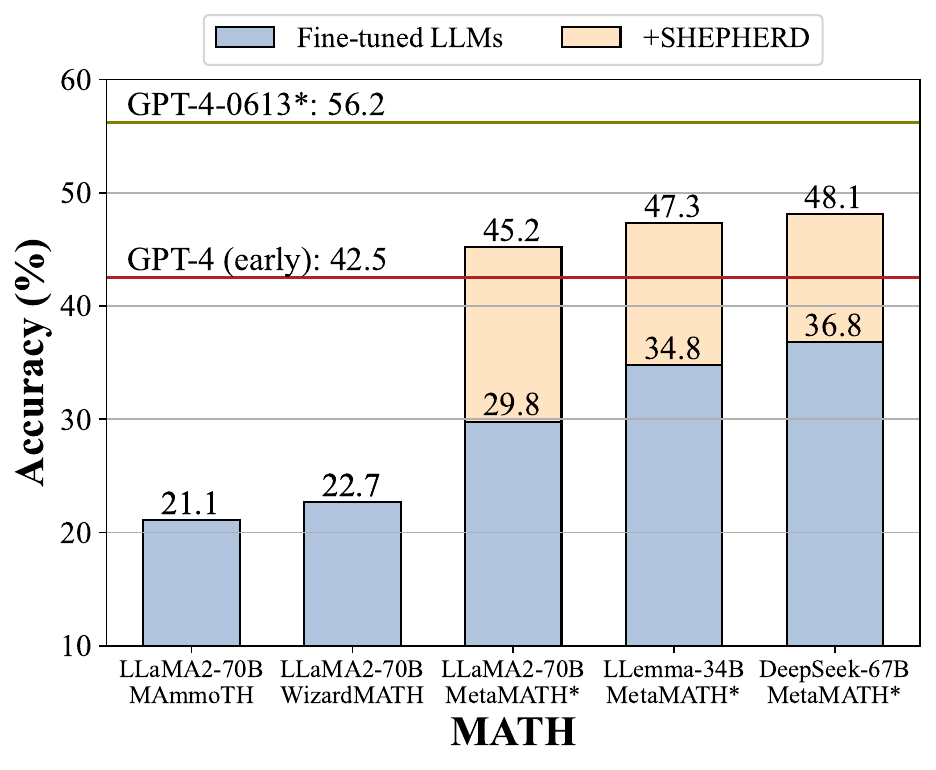}
}
\vspace{-0.1in}
\caption{We evaluate the performance of various LLMs with \methodname on the GSM8K and MATH datasets. All base models are finetuned with the MetaMath dataset \citep{yu2023metamath}. The +SHEPHERD results are obtained by selecting the best one from 256 candidates using \methodname. We observe that \methodname is compatible with different LLMs. The results of GPT-4 (early) are from \cite{bubeck2023sparks}.}
\label{fig:number-candidate}
\end{figure}

\section{Introduction}

Large language models (LLMs) have demonstrated remarkable capabilities across various tasks \citep{park2023generative,kaddour2023challenges,song2306restgpt,li2023m,wang2023voyager,chen2023towards,zheng2023judging,wang2023large}, 
However, even the most advanced LLMs face challenges in complex multi-step mathematical reasoning problems \citep{lightman2023let,huang2023large}. 
To address this issue, prior research has explored different methodologies, such as pre-training \citep{azerbayev2023llemma}, fine-tuning \citep{luo2023wizardmath,yu2023metamath,wang2023making}, prompting \citep{cot,fu2022complexity}, and verification \citep{self-consistency,li-etal-2023-making,zhu-etal-2023-solving,leviathan2023fast}.
Among these techniques, verification has recently emerged as a favored method.
The motivation behind verification is that relying solely on the top-1 result may not always produce reliable outcomes. A verification model can rerank candidate responses, ensuring higher accuracy and consistency in the outputs of LLMs.
In addition, a good verification model can also offer invaluable feedback for further improvement of LLMs \citep{uesato2022solving,wang2023making,pan2023let}.

The verification models generally fall into the outcome reward model (ORM) \citep{cobbe2021training,ovm} and process reward model (PRM) \citep{li-etal-2023-making,uesato2022solving,lightman2023let,ma2023let}.  The ORM assigns a confidence score based on the entire generation sequence, whereas the PRM evaluates the reasoning path step-by-step.  PRM is advantageous due to several compelling reasons.  One major benefit is its ability to offer precise feedback by identifying the specific location of any errors that may arise, which is a valuable signal in reinforcement learning and automatic correction. Besides, The PRM exhibits similarities to human behavior when assessing a reasoning problem. If any steps contain an error, the final result is more likely to be incorrect, mirroring the way human judgment works. 
However, gathering data to train a PRM can be an arduous process. \cite{uesato2022solving} and \cite{lightman2023let} utilize human annotators to provide process supervision annotations, enhancing the performance of PRM. 
Nevertheless, annotation by humans, particularly for intricate multi-step reasoning tasks that require advanced annotator skills, can be quite costly, which hinders the advancement and practical application of PRM.

To tackle the problem, in this paper, we propose an automatic process annotation framework. Inspired by Monte Carlo Tree Search \citep{kocsis2006bandit,coulom2006efficient,silver2016mastering,swiechowski2023monte}, we define the quality of an intermediate step as its potential to deduce the correct final answer. 
By leveraging the correctness of the answer, we can automatically gather step-wise supervision.  
Specifically, given a math problem with a golden answer and a step-by-step solution, to achieve the label of a specific step, we utilize a fine-tuned LLM to decode multiple subsequent reasoning paths from this step. 
We further validate whether the decoded final answer matches with the golden answer. 
If a reasoning step can deduce more correct answers than another, it would be assigned a higher correctness score.

We use this automatic way to construct the training data for \methodname, and verify our ideas on two widely used mathematical benchmarks, GSM8K~\citep{cobbe2021training} and MATH~\citep{MATH}.
We explore the effectiveness of \methodname in two scenarios:
1) verification:  \methodname is utilized for reranking multiple outputs generated by LLMs;
2) reinforcement learning: \methodname is employed to reinforce LLMs with step-by-step Proximal Policy Optimization (PPO).
With the verification of \methodname, a series of open-source LLMs from 7B to 70B demonstrates exceptional performance. 
For instance, the step-by-step PPO with \methodname significantly improves the accuracy of Mistral-7B (77.9\%$\to$84.1\% on GSM8K and 28.6\%$\to$33.0\% on MATH).
The accuracy can be further enhanced to 89.1\% and 43.5\% on GSM8K and MATH with verification.
DeepSeek 67B \citep{DeepSeek-llm} achieves accuracy rates of 93.3\% on the GSM8K dataset and 48.1\% on the MATH dataset with verification of \methodname.
To the best of our knowledge, these results are unprecedented for open-source models that do not rely on additional tools.

Our main contributions are as follows: 

1) We propose a framework to automatically construct process supervision datasets without human annotations for math reasoning tasks. 

2) We evaluate our method on both step-by-step verification and reinforcement learning scenarios. Extensive experiments on two widely used mathematical benchmarks - GSM8K and MATH, in addition to a series of LLMs ranging from 7B to 70B, demonstrate the effectiveness of our method.

3) We empirically analyze the key factors for training high-performing process reward models, shedding light on future directions toward improving reasoning capability with automatic step-by-step verification and supervision.

\section{Related Works}
\paragraph{Improving and eliciting mathematical reasoning abilities of LLMs.}
Mathematical reasoning tasks are one of the most challenging tasks for LLMs. 
Researchers have proposed various methods to improve or elicit the mathematical reasoning ability of LLMs, which can be broadly divided into three groups:
1) \textit{pre-training}: The pre-training methods \citep{gpt4,palm,LLaMA,azerbayev2023llemma} pre-train LLMs on a vast of datasets that are related to math problems, such as the Proof-Pile and ArXiv \citep{azerbayev2023llemma} with a simple next token prediction objective. 
% These studies suggest that a larger model pre-trained on more data tends to have a better mathematical reasoning ability;
2) \textit{fine-tuning}: The fine-tuning methods \citep{yu2023metamath,luo2023wizardmath,yue2023mammoth,wang2023making,gou2023tora} can also enhance the mathematical reasoning ability of LLMs. 
The core of fine-tuning usually lies in constructing high-quality question-response pair datasets with a chain-of-thought reasoning process. 
% Training LLMs on such chain-of-thought datasets can further improve the mathematical reasoning performance of pre-trained LLMs;
and 3) \textit{prompting}: The prompting methods \citep{cot,zhang2023cumulative,fu2022complexity,bi2023program} aim to elicit the mathematical reasoning ability of LLMs by designing prompting strategy without updating the model parameters, which is very convenient and practical.

\paragraph{Mathematical reasoning verification for LLMs.} Except for directly improving and eliciting the mathematical reasoning potential of LLMs, 
the reasoning results can be boosted via an extra verifier for selecting the best answer from multiple decoded candidates.
There are two primary types of verifiers: the Outcome Reward Model (ORM) and the Process Reward Model (PRM). The ORM allocates a score to the entire solution while the PRM assigns a score to each individual step in the reasoning process.
Recent findings by \citep{lightman2023let} suggest that PRM outperforms ORM.
In addition to verification, reward models can offer invaluable feedback for further training of generators \citep{uesato2022solving,pan2023let}.
Compared to ORM, PRM provides more detailed feedback, demonstrating greater potential to enhance generator \citep{wu2023fine}. 
However, training a PRM requires access to expensive human-annotated datasets \citep{uesato2022solving,lightman2023let}, which hinders the advancement and practical application of PRM.
Therefore, in this paper, we aim to build a PRM for mathematical reasoning without human annotation, and we explore the effectiveness of the automatic PRM with both verification and reinforcement learning scenarios.

\section{Methodology}
In this section, we first present our task formulation to evaluate the performance of reward models (\S\ref{subsec:task_formulation}).
Subsequently, we outline two typical categories of reward models, ORM and PRM(\S\ref{subsec:reward_model}).
Then, we introduce our methodology to automatically build the training dataset for PRM(\S\ref{subsec:apa}), breaking the bottleneck of heavy reliance on manual annotation in existing
work \citep{uesato2022solving,lightman2023let}.

\subsection{Task Formulation}
\label{subsec:task_formulation}
We evaluate the performance of the reward model in two scenarios:
\paragraph{Verification} Following \citep{lightman2023let}, we consider a best-of-N selection evaluation paradigm.
Specifically, given a problem $p$ in the testing set, we sample N candidate solutions from a generator. These candidates are then scored using a reward model, and the highest-scoring solution is selected as the final answer. 
An enhanced reward model elevates the likelihood of selecting the solution containing the correct answer, consequently raising the success rate in solving mathematical problems for LLMs.
\paragraph{Reinforcement learning} We also use the automatically constructed PRM to supervise LLMs with step-by-step PPO. In this scenario, we evaluate the accuracy of the LLMs' greedy decoding output. 
An enhanced reward model is instrumental in training higher-performing LLMs.

\subsection{Reward Models for Mathematical Problem}
\label{subsec:reward_model}
\paragraph{ORM} Given a mathematical problem $p$ and its solution $s$, ORM ($P \times S \to \mathbb{R}$) assigns a single real-value to $s$ to indicate whether $s$ is correct.
ORM is usually trained with a cross-entropy loss \citep{cobbe2021training, li-etal-2023-making}:
\begin{equation}
    \mathcal{L}_{ORM} = y_s \log r_s + (1 - y_s) \log (1 - r_s),
\end{equation}
where $y_s$ is the golden answer of the solution $s$, $y_s=1$ if $s$ is correct, otherwise $y_s=0$. $r_s$ is the sigmoid score of $s$ assigned by ORM. 
The success of the reward model hinges on the effective construction of the high-quality training dataset. As the math problem usually has a certain answer, we can automatically construct the training set of ORM by two steps: 1) sampling some candidate solutions for a problem from a generator; 2) assigning the label to each sampling solution by checking whether its answer is correct. 
Although \textit{false positives} solutions that reach the correct answer with incorrect reasoning will be misgraded, previous studies have proven that it is still effective for training a good ORM \citep{lightman2023let,ovm}.

\paragraph{PRM} Take a step further, PRM ($P \times S \to \mathbb{R}^{+}$) assigns a score to each reasoning step of $s$, which is usually trained with:
\begin{equation}
    \mathcal{L}_{PRM} = \sum_{i=1}^K y_{s_i} \log r_{s_i}+ (1 - y_{s_i}) \log (1 - r_{s_i}),
    \label{eq:prm}
\end{equation}
where $y_{s_i}$ is the golden answer of $s_i$ (the $i$-th step of $s$), $r_{s_i}$ is the sigmoid score of $s_i$ assigned by PRM and $K$ is the number of reasoning steps for $s$. \citep{lightman2023let} also conceptualizes the PRM training as a three-class classification problem, in which each step is classified as either `good', `neutral', or `bad'. In this paper, we found that there is not much difference between the binary and the three classifications, and we regard PRM training as the binary classification.
Compared to ORM, PRM can provide more detailed and reliable feedback \citep{lightman2023let}. 
However, there are currently no automated methods available for constructing high-quality PRM training datasets.
Previous works \citep{uesato2022solving,lightman2023let} typically resort to costly human annotations. While PRM manages to outperform ORM \citep{lightman2023let}, the annotation cost invariably impedes both the development and application of PRM.

\begin{figure*}[t]
    \centering
    \includegraphics[width=1\linewidth]{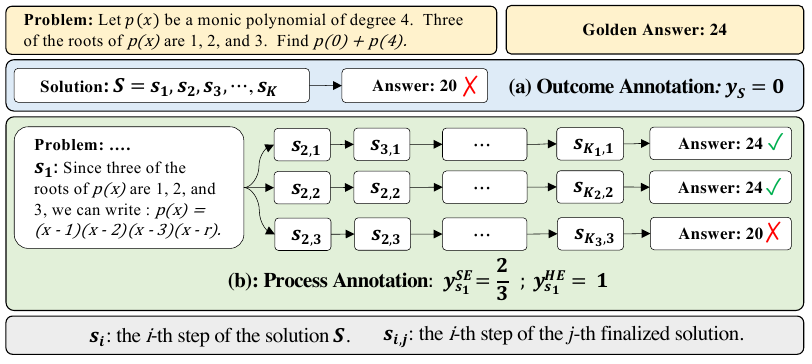}
    \caption{Comparison for previous automatic outcome annotation and our automatic process annotation. (a): automatic outcome annotation assigns a label to the entire solution $S$, dependent on the correctness of the answer; (b) automatic process annotation employs a `completer' to finalize N reasoning processes (N=3 in this figure) for an intermediate step ($s_1$ in this figure), subsequently use hard estimation (HE) and soft estimation (SE) to annotate this step based on all decoded answers.}
    \label{fig:overview}
\end{figure*}

\subsection{Automatic Process Annotation}
\label{subsec:apa}
In this section, we propose an automatic process annotation framework to mitigate the annotation cost issues associated with PRM. We first define the quality of a reasoning step, followed by the introduction of our solution that obviates the necessity for human annotation.

\subsubsection{Definition} Inspired by Monto Carlo Tree Search \citep{kocsis2006bandit,coulom2006efficient,silver2016mastering,swiechowski2023monte}, we define the quality of a reasoning step as \textit{its potential to deduce the correct answer.}
This criterion stems from the primary objective of the reasoning process, which essentially is a cognitive procedure aiding humans or intelligent agents in reaching a well-founded outcome \citep{huang-chang-2023-towards}. Therefore, a step that has the potential to deduce a well-founded result can be considered a good reasoning step.  Analogous to ORM, this definition also introduces some degree of noise. Nevertheless, we find that it is beneficial for effectively training a good PRM.

\subsubsection{Solution}

\paragraph{Completion} To quantify and estimate the \textit{potential} for a give reasoning step $s_i$, 
as shown in Figure \ref{fig:overview},
we use a `completer' to finalize N subsequent reasoning processes from this step: $\{(s_{i+1,j}, \cdots, s_{K_j,j}, a_j)\}_{j=1}^N$, where $a_j$ and $K_j$ are the decoded answer and the total number of steps for the $j$-th finalized solution, respectively.
Then, we estimate the potential of this step based on the correctness of all decoded answers $A=\{a_j\}_{j=1}^N$. 

\paragraph{Estimation} In this paper, we use two methods to estimate the quality $y_{s_i}$ for the step $s_i$, hard estimation (HE) and soft estimation (SE). HE supposes that a reasoning step is good as long as it can reach the correct answer $a^*$:
\begin{equation}
y_{s_i}^{HE}=\left\{
\begin{aligned}
1 & & \exists a_j \in A, a_j = a^*\\
0 & & \mathrm{Otherwise} \\
\end{aligned}
\right.
\end{equation}
SE assumes the quality of a step as the frequency with which it reaches the correct answer:
\begin{equation}
    y_{s_i}^{SE} = \frac{\sum_{j=1}^N\mathbb{I} (a_j = a^*)}{N}.
\end{equation}
Once we gather the label of each step, we can train PRM with the cross-entropy loss.
In conclusion, our automatic process annotation framework defines the quality of a step as its potential to deduce the correct answer and achieve the label of each step by completion and estimation.

\subsection{Ranking for Verification}
Following \citep{lightman2023let}, we use the minimum score across all steps to represent the final score of a solution assigned by PRM. We also explore the combination of self-consistency and reward models following \citep{li-etal-2023-making}. In this context, we initially classify solutions into distinct groups according to their final answers. Following that, we compute the aggregate score for each group.
Formally, the final prediction answer based on N candidate solutions is:
\begin{equation}
    a_{sc+rm} = \argmax_{a} \sum_{i=1}^N \mathbb{I}(a_i = a) \cdot RM(p, S_i).
\end{equation}
Where $RM(p, S_i)$ is the score of the $i$-th solution assigned by ORM or PRM for problem $p$.

\subsection{Reinforce Learning with Process Supervision}
Upon achieving PRM, we employ reinforcement learning to train LLMs. We implement Proximal Policy Optimization (PPO) in a step-by-step manner. 
This method differs from the conventional strategy that utilizes PPO with ORM, which only offers a reward at the end of the response. Conversely, our step-by-step PPO offers rewards at the end of each reasoning step.

% \begin{equation}
% r_t=\left\{
% \begin{aligned}
% PRM(p, S_i),  & & t \mathrm{ \ is \ the \ end \ index \ of \ the \ step \ S_i} \\
% 0 & & \mathrm{Otherwise} \\
% \end{aligned}
% \right.
% \end{equation}

\section{Experiments}
% \subsection{Experimental Setups}
\paragraph{Datasets}
We conduct our experiments using two widely used math reasoning datasets, GSM8K \citep{cobbe2021training} and MATH \citep{MATH}. 
For the GSM8K dataset, we leverage the whole test set in both verification and reinforcement learning scenarios. 
For the MATH dataset, in the verification scenario,
due to the computation cost,  we employ a subset MATH500 that is identical to the test set of \citet{lightman2023let}. The subset consists of 500 representative problems, and we find that the subset evaluation produces similar results to the full-set evaluation.
To assess different verification methods, we generate 256 candidate solutions for each test problem.
We report the mean accuracy of $3$ groups of sampling results.
In the reinforcement learning scenario, we use the whole test set to evaluate the model performance.
We train LLMs with MetaMATH \citep{yu2023metamath}.

\begin{table*}[t]
    \centering
    \small
    \scalebox{1}{
    \begin{tabular}{llcc}
    \toprule
     \textbf{Models}  & \textbf{Verifiers} & \textbf{GSM8K}  & \textbf{MATH500}  \\
     \midrule
    \multirow{5}{*}{\rotatebox{0}{LLaMA2-70B: MetaMATH}} &  Self-Consistency   & 88.0 & 39.4 \\
     & ORM  & 91.8 & 40.4  \\
    & Self-Consistency + ORM  & 92.0 & 42.0  \\
    \cmidrule(r){2-4}
    & \methodname (Ours) & {93.2} & 44.5 \\
    & Self-Consistency + \methodname (Ours) & 92.4 &  {45.2}  \\
    \midrule
    \multirow{5}{*}{\rotatebox{0}{LLemma-34B: MetaMATH}} &  Self-Consistency & 82.6 & 44.2  \\
    & ORM  & 90.0 & 43.7 \\
    & Self-Consistency + ORM  & 89.6 & 45.4  \\
    \cmidrule(r){2-4}
    & \methodname (Ours) & {90.9} & 46.0  \\
    & Self-Consistency + \methodname (Ours) & 89.7 & {47.3} \\
    \midrule
    \multirow{5}{*}{\rotatebox{0}{DeepSeek-67B: MetaMATH}} & Self-Consistency  & 88.2 & 45.4  \\
     & ORM  & 92.6 &  45.3 \\
    & Self-Consistency + ORM  & 92.4  & 47.0  \\
    \cmidrule(r){2-4}
    & \methodname (Ours) & \textbf{93.3} & 47.0  \\
    & Self-Consistency + \methodname (Ours) &  92.5 & \textbf{48.1}   \\
    \bottomrule
    \end{tabular}}
    \caption{Performances of different LLMs on GSM8K and MATH with different verification strategies. The reward models are trained based on LLama2-70B and LLemma-34B on GSM8K and MATH, respectively. The verification is based on 256 outputs.}
    \label{tab:main}
    % \vspace{-0.1in}
\end{table*}

\paragraph{Parameter Setting}
\label{sec:param}
Our experiments are based on a series of large language models, LLaMA2-7B/13B/70B \citep{LLaMA}, LLemma-7B/34B \citep{azerbayev2023llemma}, Mistral-7B \citep{jiang2023mistral} and DeepSeek-67B \citep{DeepSeek-llm}. 
We train the generator and completer for 3 epochs on MetaMATH. 
We train the Mistral-7B with a learning rate of 5e-6. 
For other models, The learning rates are set to 2e-5, 1e-5, and 6e-6 for the 7B/13B, 34B, and 67B/70B LLMs, respectively.
To construct the training dataset of ORM and PRM, we train 7B and 13B models for a single epoch on the GSM8K and MATH training sets. Subsequently, we sample 15 solutions per problem from each model for the training set. Following this, we eliminate duplicate solutions and annotate the solutions at each step. 
We use LLemma-7B as the completer with the decoded number N=8.
Consequently, we obtain around 170k solutions for GSM8K and 270k solutions for MATH.
For verification,
we choose LLaMA2-70B and LLemma-34B as the base models to train reward models for GSM8K and MATH, respectively. 
For reinforcement learning, 
we choose Mistral-7B as the base model to train reward models and use it to supervise LLama2-7B and Mistral-7B generators.
The reward model is trained in 1 epoch with a learning rate 1e-6.
For the sake of convenience, we train the PRM using the hard estimation version because it allows us to utilize a standard language modeling pipeline by selecting two special tokens to represent `has potential' and `no potential' labels, thereby eliminating the need for any specific model adjustments.
In reinforcement learning, the learning rate is 4e-7 and 1e-7 for LLaMA2-7B and Mistral-7B, respectively.
The Kullback-Leibler coefficient is set to 0.04.
We implement a cosine learning rate scheduler, employing a minimal learning rate set to 1e-8. 
We use 3D parallelism provided by hfai\footnote{\url{https://doc.hfai.high-flyer.cn/index.html}} to train all models with the max sequence length of 512. 

\paragraph{Baselines and Metrics}
In the verification scenario, following \citep{lightman2023let}, we evaluate the performance of our reward model by comparing it against the Self-consistency (majority voting) and outcome reward model. The accuracy of the best-of-N solution is utilized as the evaluation metric.
For PRM, the minimum score across all steps is adopted to represent the final score of a solution. 
In the reinforcement scenario, we compare our step-by-step supervision with the outcome supervision provided by ORM, and Rejective Sampling Fine-tuning (RFT) \citep{yuan2023scaling}, we sample 8 responses for each question in MetaMATH for RFT.
We use the accuracy of LLMs' greedy decoding output to assess the performance.

\subsection{Main Results}
\paragraph{\methodname as verifier}
Table \ref{tab:main} presents the performance comparison of various methods on GSM8K and MATH. 
We find that:
1) As the verifier, \methodname consistently outperforms self-consistency and ORM on two datasets with all generators. Specifically, enhanced by \methodname, DeepSeek-67B achieves 93.3\% and 48.1\% accuracy on GSM8K and MATH;
2) In comparison to GSM8K, PRM achieves a greater advantage over ORM on the more challenging MATH dataset;
This outcome aligns with the findings in \cite{uesato2022solving} and \cite{lightman2023let}. The former discovers that PRM and ORM yield similar results on GSM8K, whereas the latter shows that PRM significantly outperforms ORM on the MATH dataset.
This could be attributed to the relative simplicity of the GSM8K dataset compared to MATH, i.e., the GSM8K dataset necessitates fewer steps for problem-solving. 
As a result, ORM operates efficiently when handling this particular dataset;
3) In GSM8K, when combined with self-consistency, there's a drop in performance, whereas in MATH, performance improves. These results indicate that if the reward model is sufficiently powerful for a task, combining it with self-consistency may harm the verification performance.

\begin{table*}[t]
    \centering
    % \small
    \scalebox{1}{
    \begin{tabular}{lcc}
    \toprule
     \textbf{Models}  &\textbf{GSM8K}  & \textbf{MATH}  \\
     \midrule
    % GPT-4 & 92.0 & 42.5  \\
    % GPT-4-0613$^*$ & 94.4 & 56.2  \\
    % \midrule
    % LLaMA2-70B: MAmmoTH \citep{yue2023mammoth} & 72.4 &  21.1 \\
    % LLaMA2-70B: WizardMATH \citep{luo2023wizardmath}  & 81.6 & 22.7  \\ 
    % LLaMA2-70B: MetaMATH \citep{yu2023metamath} & 82.3 & 26.6\\
    % LLemma-34B: MetaMATH$^*$ & 75.8 & - \\
    % DeepSeek-67B: MetaMATH$^*$ & 82.8 & - \\
    % \midrule
    % \midrule
    LLaMA2-7B: MetaMATH  & 66.6 & 19.2   \\
    \midrule
    \quad + RFT & 68.5 & 19.9  \\
    \quad +  ORM-PPO  &  70.8 & 20.8  \\
    \midrule
    \quad + \methodname-step-by-step-PPO (Ours)   & 73.2 & 21.6 \\
    \midrule
    \midrule
    Mistral-7B: MetaMATH & 77.9 & 28.6 \\
    \midrule
    \quad + RFT & 79.0 & 29.9 \\
    \quad +  ORM-PPO & 81.8 & 31.3 \\
    \midrule
    \quad + \methodname-step-by-step-PPO (Ours)   & \textbf{84.1} & \textbf{33.0} \\
    \bottomrule
    \end{tabular}}
    \caption{Performances of different 7B models on GSM8K and MATH with greedy decoding. We use the questions in MetaMATH for RFT and PPO training. Both LLaMA2-7B and Mistral-7B are supervised by Mistral-7B-ORM and -\methodname.}
    \label{tab:main-rl}
    % \vspace{-0.1in}
\end{table*}

\paragraph{\methodname as reward model on reinforcement learning}
Table \ref{tab:main-rl} presents the performance of different LLMs with greedy decoding outputs.
As is shown:
1) step-by-step PPO significantly improves the performance of two supervised fine-tuned models.
For example, Mistral-7B with step-by-step PPO achieves 84.1\% and 33.0\% on the GSM8K and MATH datasets, respectively;
2) RFT only slightly improves the model performance, we believe this is because MetaMATH already has conducted some data augmentation strategies like RFT;
3) the vanilla PPO with ORM  can also enhance the model performance.
However, it does not perform as well as the step-by-step PPO supervised by \methodname, demonstrating the potential of step-by-step supervision.

\paragraph{\methodname as both reward models and verifiers}
We also combine the reinforcement learning and the verification.
As shown in Table \ref{tab:main-rl+ve}:
1) reinforcement learning and verification are complementary. For example, in MATH, step-by-step PPO Mistral-7B outperforms supervised fine-tuning Mistral-7B 7.2\% accuracy with self-consistency as the verifier;
The performance gap is even larger than that of greedy decoding results, i.e., 4.4\%;
2) after reinforcement learning, the vanilla verification methods with only reward models is inferior to self-consistency, we think the reason is that the initial reward model is not sufficient to supervise the more powerful model after PPO.
These results can also show the potential of iterative reinforcement learning, which we leave for future work.

\section{Analysis}

\subsection{Performance with Different Number of Candidate Solutions}
Figure \ref{fig:number-candidate} illustrates the performance comparison of various strategies when applied to different numbers of candidates ranging from 1 to 256 on two benchmarks. The key observations are as follows:
1) PRM exhibits consistent superior performance when compared to both ORM and majority voting, with the degree of this superiority becoming more pronounced as N escalates.
2) In MATH, our automatically annotated datasets outperform the human-annotated PRM800K \citep{lightman2023let}. We ascribe this superiority to the distribution gap and the data quantity. Specifically, PRM800K is annotated based on the output from GPT-4, and consequently, a discrepancy arises for the output of open-source LLaMA models fine-tuned on MetaMATH. Furthermore, when considering the quantity of data, our automated reward model data exhibits both high scalability and a reduced labeling cost. Consequently, our dataset is four times larger than that provided in PRM800K.
Overall, these results further underscore the effectiveness and potential of our method.

\begin{table*}[t]
    \centering
    \small
    \scalebox{1}{
    \begin{tabular}{llcc}
    \toprule
     \textbf{Models}  & \textbf{Verifiers} & \textbf{GSM8K}  & \textbf{MATH500}  \\
     \midrule
    \multirow{5}{*}{\rotatebox{0}{Mistral-7B: MetaMATH}} &  Self-Consistency   & 83.9 & 35.1\\
     & ORM  & 86.2 & 36.4  \\
    & Self-Consistency + ORM  & 86.6 & 38.0  \\
    \cmidrule(r){2-4}
    & \methodname (Ours) & 87.1 & 37.3 \\
    & Self-Consistency + \methodname (Ours) & 86.3  &  38.3  \\
    \midrule
    \multirow{2}{*}{\rotatebox{0}{Mistral-7B: MetaMATH}} &  Self-Consistency & 87.4 & 42.3  \\
    & ORM  & 87.6 &  41.3 \\
    \multirow{2}{*}{\rotatebox{0}{+step-by-step PPO (Ours)}} & Self-Consistency + ORM  & 89.0  & 43.1  \\
    \cmidrule(r){2-4}
    & \methodname (Ours) & 88.4 & 41.1  \\
    & Self-Consistency + \methodname (Ours) & \textbf{89.1} & \textbf{43.5}  \\
    \bottomrule
    \end{tabular}}
    \caption{Results of reinforcement learning and verification combination. The reward models are trained based on Mistral-7B. The verification is based on 256 outputs.}
    \label{tab:main-rl+ve}
    % \vspace{-0.1in}
\end{table*}

\subsection{Quality of the Automatic Process Annotations}

In this section, we explore the quality of our automatic PRM dataset.
To achieve this, we manually annotate $160$ steps sampled from the training set of GSM8K and use different completers to infer from each step to achieve their label. We find that:
\begin{figure}[t]
\centering
\subfigure{
\includegraphics[width=0.48\textwidth]{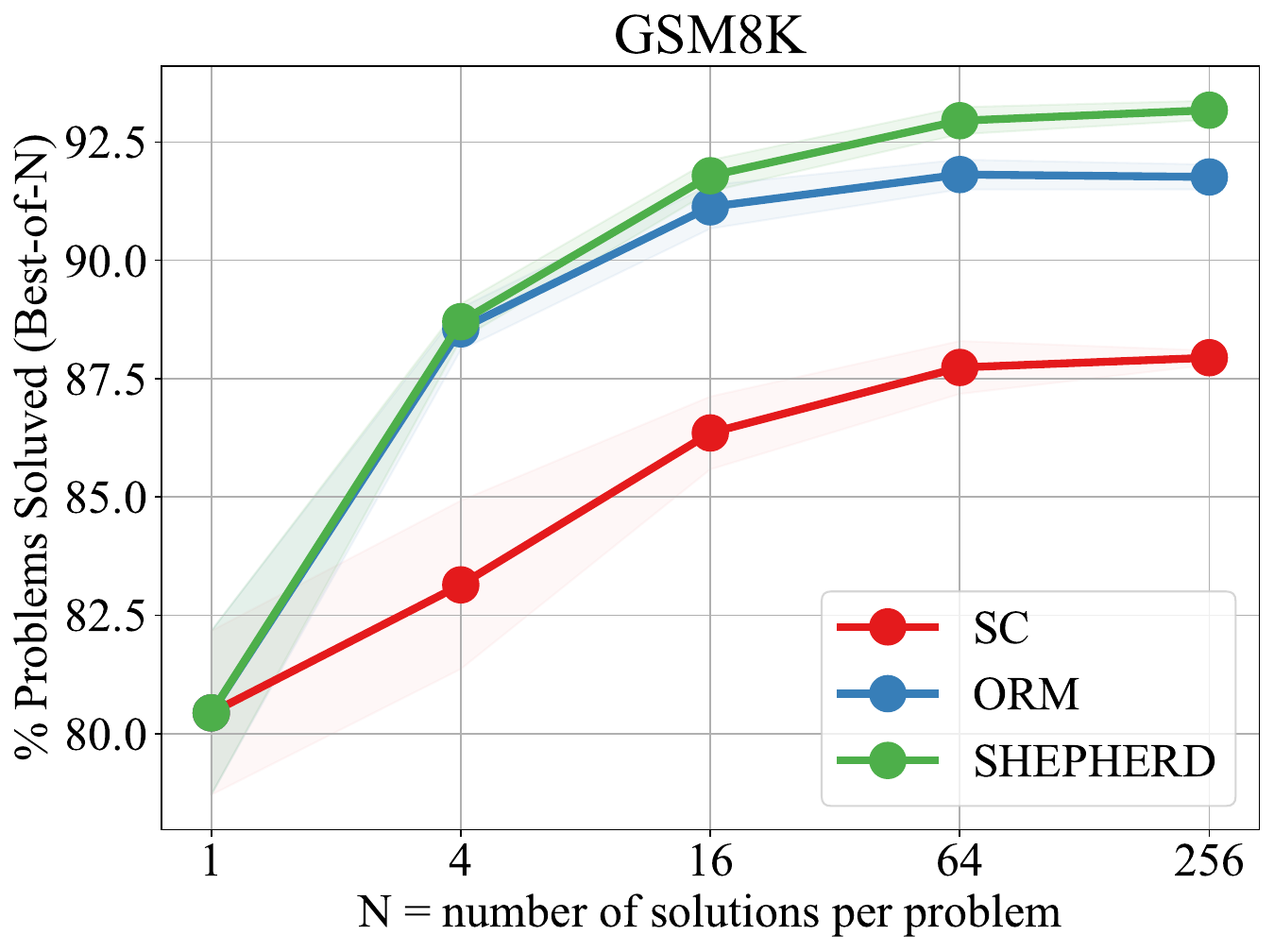}
}
\subfigure{
\includegraphics[width=0.475\textwidth]{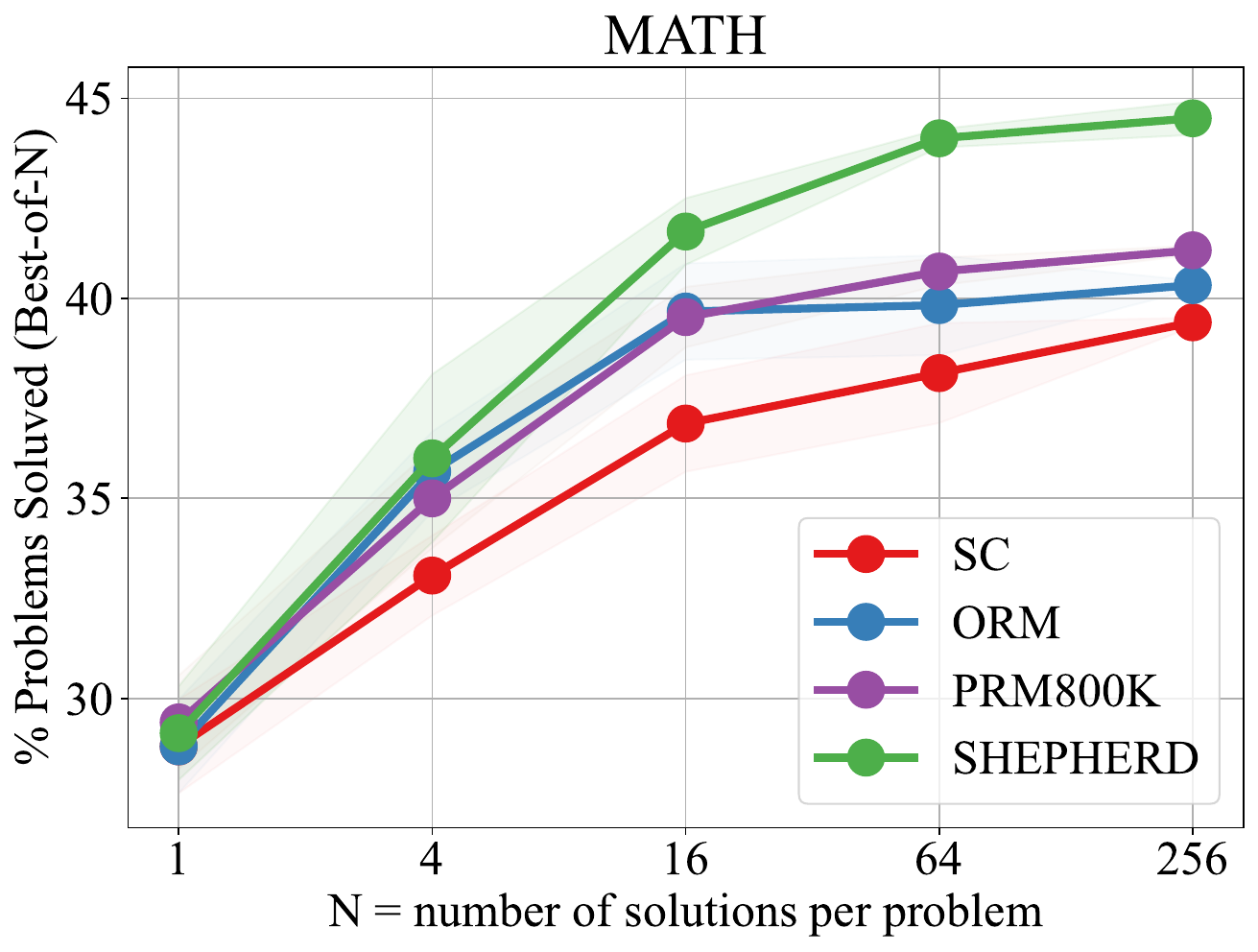}
}
\caption{Performance of LLaMA2-70B using different verification strategies across different numbers of solution candidates on GSM8K and MATH.}
\label{fig:number-candidate}
\end{figure}

\begin{figure}[t]
\centering
\subfigure{
\includegraphics[width=0.33\textwidth]{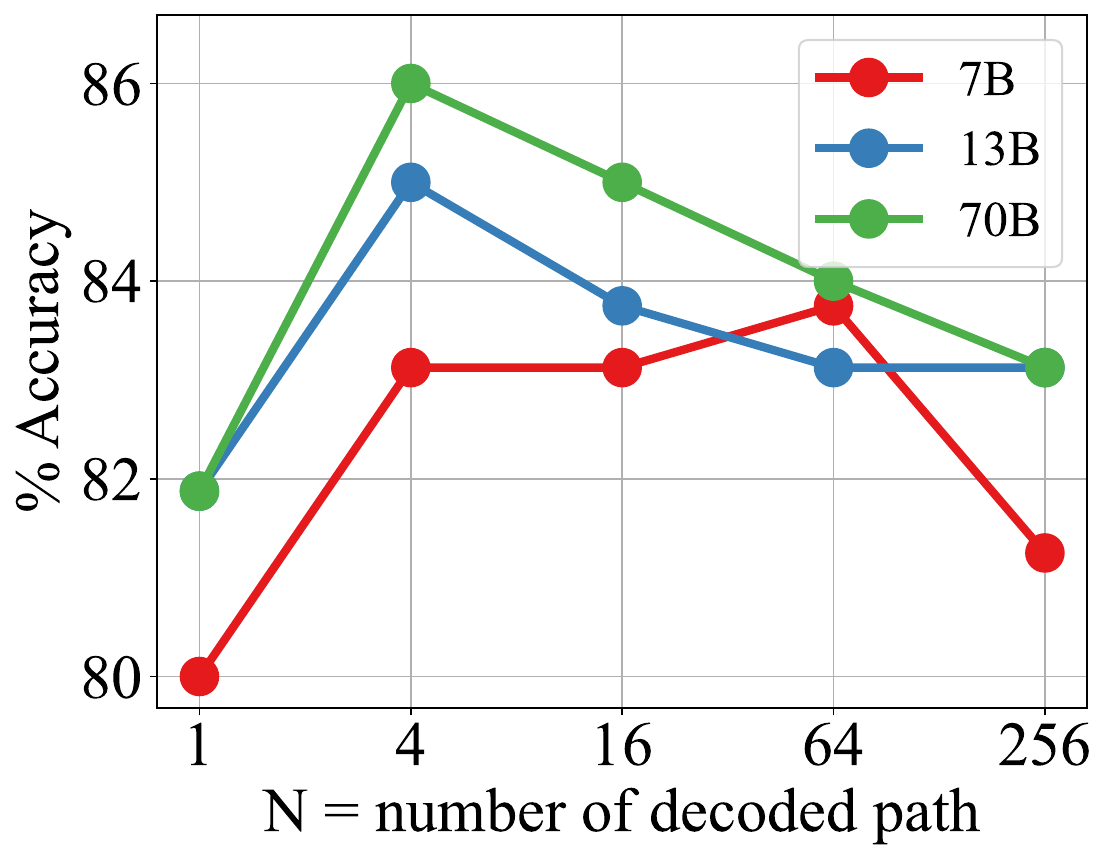}\hspace{-3mm}
}
\subfigure{
\includegraphics[width=0.33\textwidth]{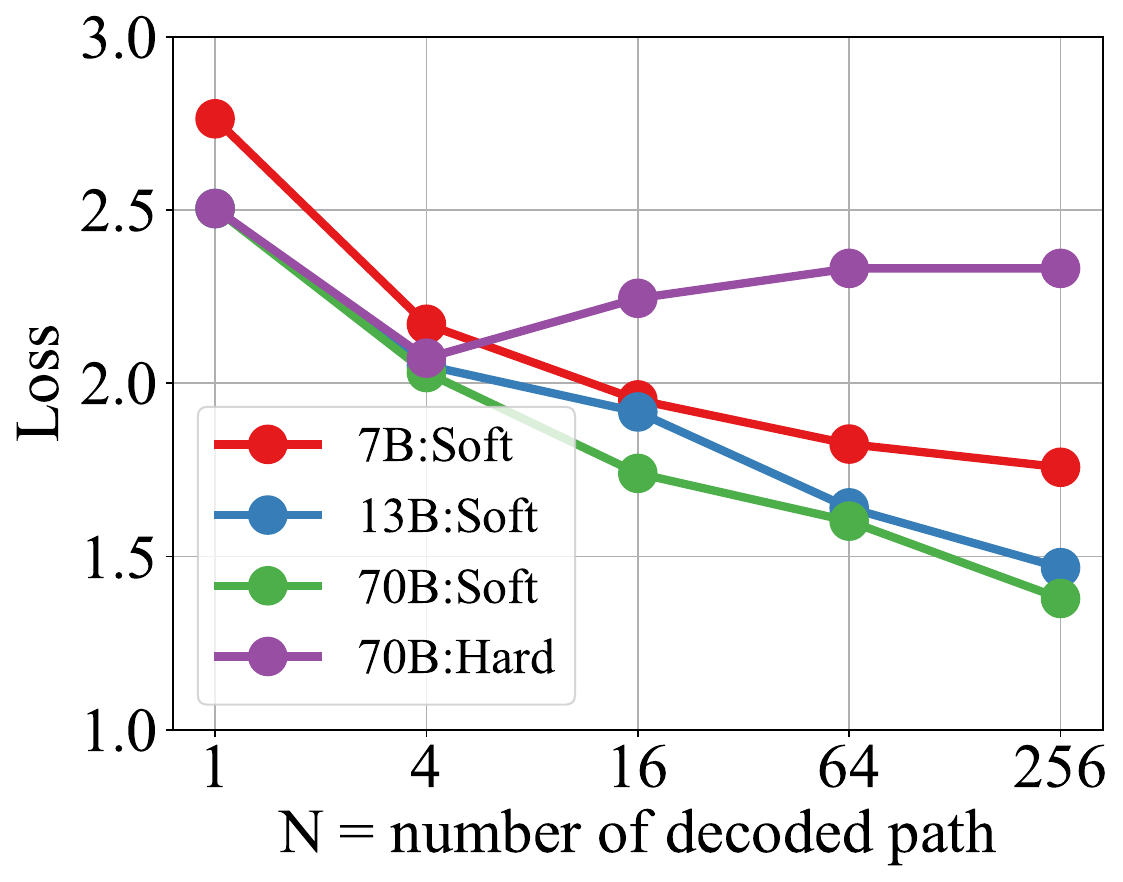}\hspace{-3mm}
}
\subfigure{
\includegraphics[width=0.315\textwidth]{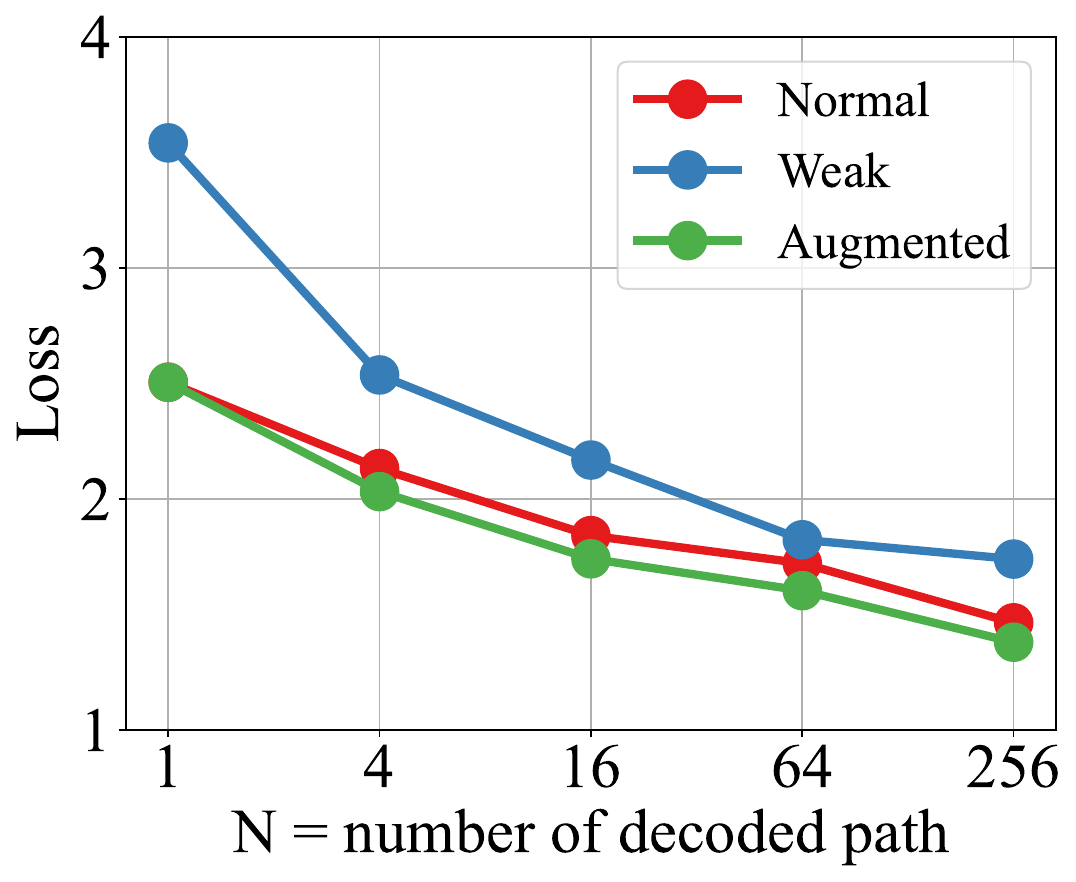}\hspace{-3mm}
}
\caption{Quality of process annotation on GSM8K. (a): Accuracy of the process annotation using different completer; (b): Loss of the process annotation using different completer; (c): Loss of the process annotation using the same completer with different training data.}
\label{fig:quality}
\end{figure}

\paragraph{Automatic process annotation exhibits satisfactory quality.} 
Figure \ref{fig:quality}(a) demonstrates that utilizing LLaMA2-70B trained on MetaMATH as the completer, the accuracy of the hard estimation (HE)  reaches 86\% when N equals 4. This suggests that our automatically constructed dataset is of high quality. However, we observed a decline in the accuracy of the constructed dataset with further increases in N. Our analysis indicates that larger values for N may lead to false positives.

Figure \ref{fig:quality}(b) shows the cross-entropy loss between SE and HE labels compared to the human-annotated distribution: as N increases, SE progressively aligns closer to the standard distribution, in contrast to HE which does not exhibit similar behavior.
It is essential to note that at N=4, HE achieves an accuracy of 86\%. We can theoretically attain higher quality data exceeding 86\%  accuracy by utilizing SE. 
However, we discovered that the performance of the verifier exhibits no substantial divergence whether trained with either SE or HE. This may be attributable to the already high-quality annotations provided by HE.

Furthermore, we also delve into other automatic process annotation methodologies. For instance, \citep{li-etal-2023-making} employs a natural language inference (NLI) model and a string match rule to annotate a given step.
The NLI-based method annotates a step as correct if it is entailment with any step in the reference solutions.
The Rule-based method annotates a step as correct if its support number precisely matches that of any steps in the reference solutions.
As demonstrated in Table \ref{tab:nli}, our annotation strategy exhibits substantial superiority over the two approaches.

\begin{table}[t]
    \centering
    \scalebox{1}{
    \begin{tabular}{llcc}
    \toprule
     \textbf{Methods} & \textbf{Models} & \textbf{Accuracy (\%)} & \textbf{Loss}   \\
     \midrule
     DIVERSE-NLI \citep{li-etal-2023-making} & DeBERTa \citep{he2020deberta} & 61.3 & 5.43 \\
     DIVERSE-NLI \citep{li-etal-2023-making} & LLaMA2-13B & 75.6 & 3.27 \\
     DIVERSE-Rule \citep{li-etal-2023-making} & - & 75.0 & 3.43 \\
     \midrule
     \methodname & LLaMA2-13B (N = 4) & 85.0  & 2.05 \\
    \bottomrule
    \end{tabular}}
    \caption{The comparison between NLI/Rule-based automatic process annotation methods from \citet{li-etal-2023-making} and our method.}
    \label{tab:nli}
\end{table}

\paragraph{The ability of the LLM completer plays an important role in the data quality.} We employ a completer to finalize multiple subsequent reasoning processes for a given step.
Therefore, we investigate the impact of the LLM completer.

Figure \ref{fig:quality}(b) presents the cross-entropy loss across diverse completers trained on MetaMath. The results indicate that a larger completer is adept at generating superior-quality datasets.
Figure \ref{fig:quality}(c) depicts the cross-entropy loss of LLaMA2-70B trained with different datasets. `Normal' denotes the original GSM8K training dataset; `Weak' refers to the Normal set excluding examples whose questions are in our 160 evaluation set; while `Augmented' symbolizes MetaMath, an augmented version of the Normal set.

The findings suggest that high-quality training sets allow the model to operate more proficiently as a completer. Importantly, the `Weak' set exhibits a markedly larger loss than other datasets. This insight drives us to infer that LLMs should acquire the questions in advance to enhance their performance as completers. 
We can also conjecture that a stronger foundational model, coupled with superior training data, could further enhance the quality of automatic annotation.

\subsection{Influence of the Pre-trained Base Models}

\begin{figure}[t]
\centering
\subfigure{
\includegraphics[width=0.253\textwidth]{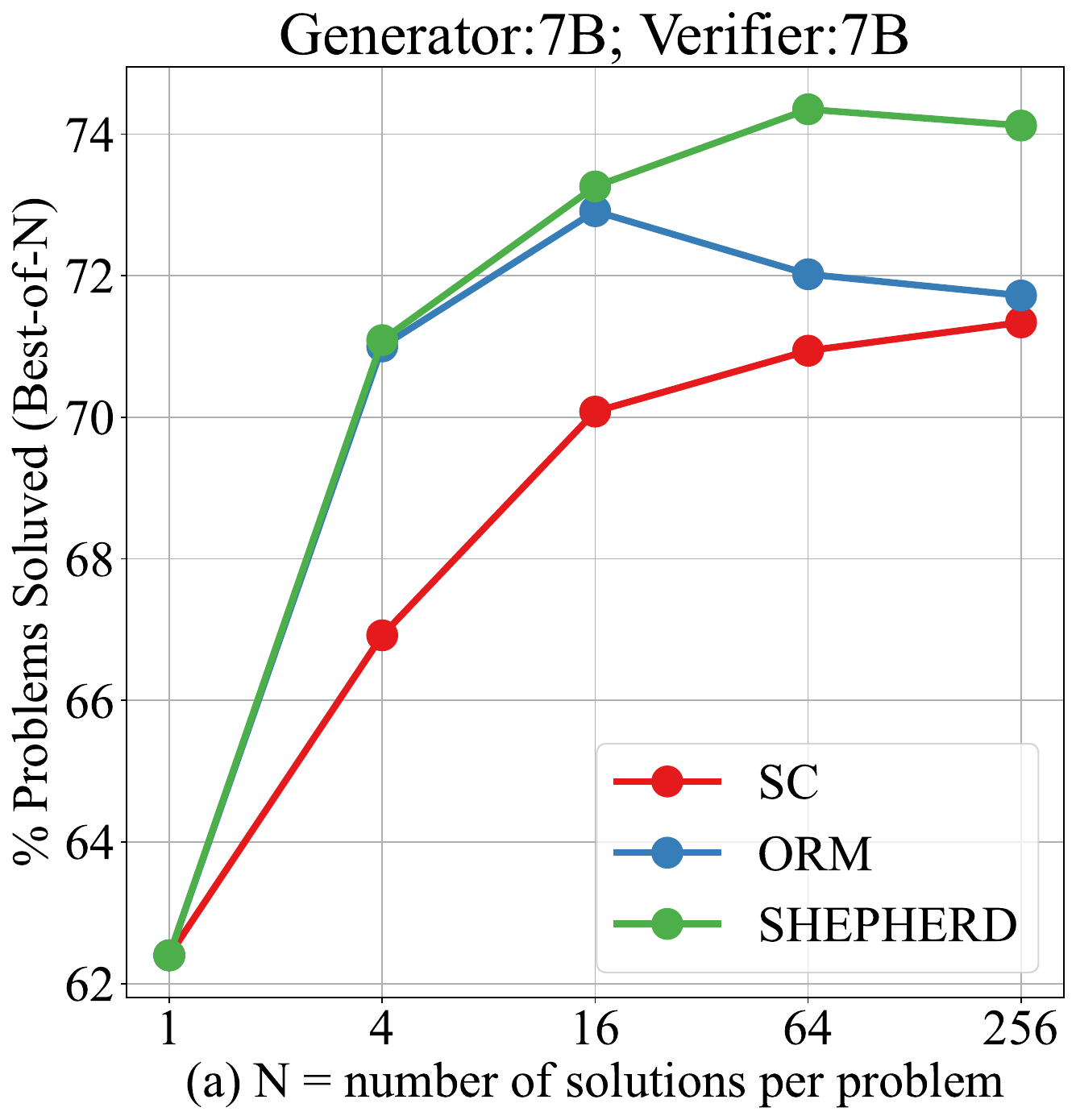}\hspace{-2mm}
}
\subfigure{
\includegraphics[width=0.24\textwidth]{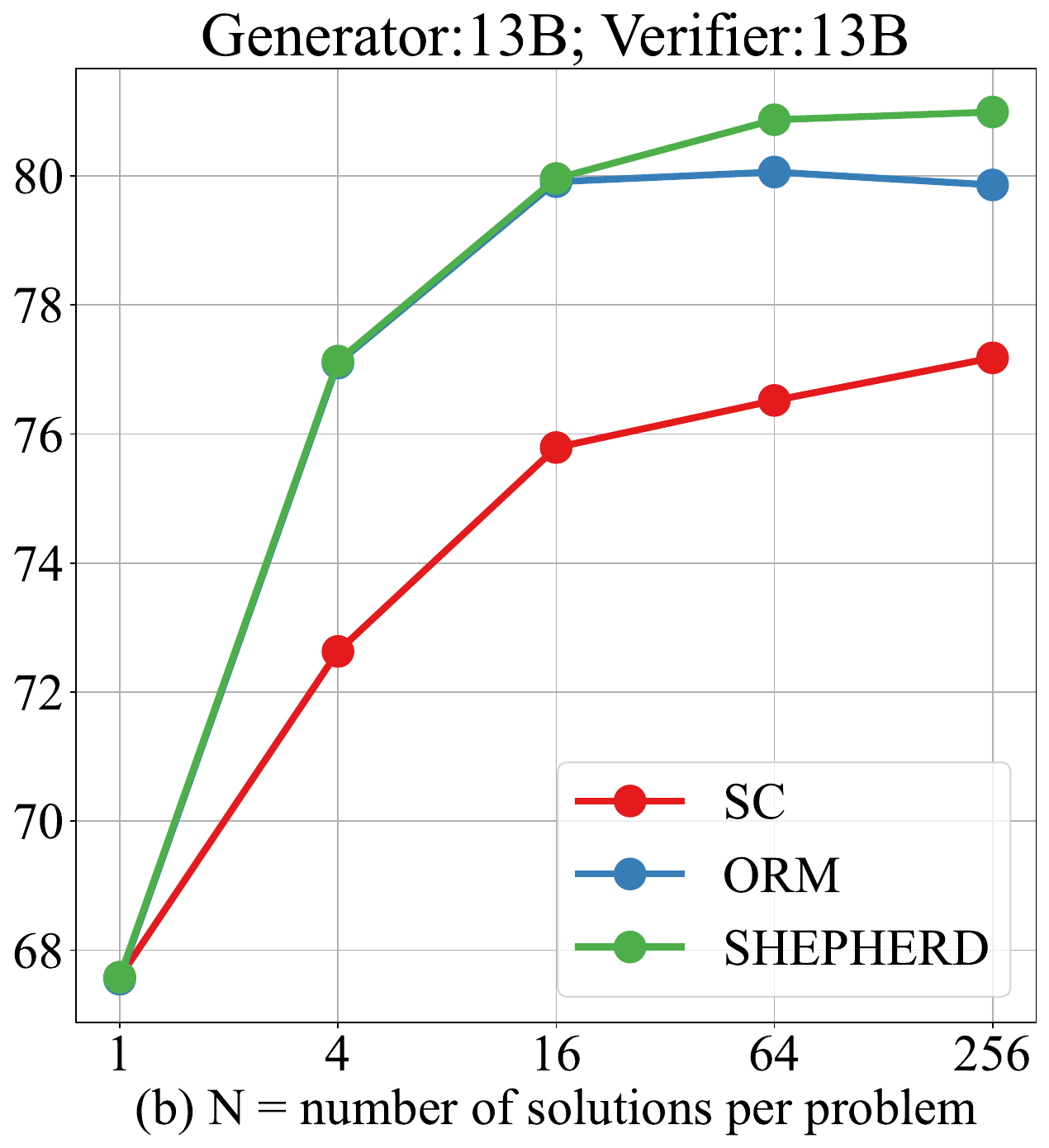}\hspace{-2mm}
}
\subfigure{
\includegraphics[width=0.24\textwidth]{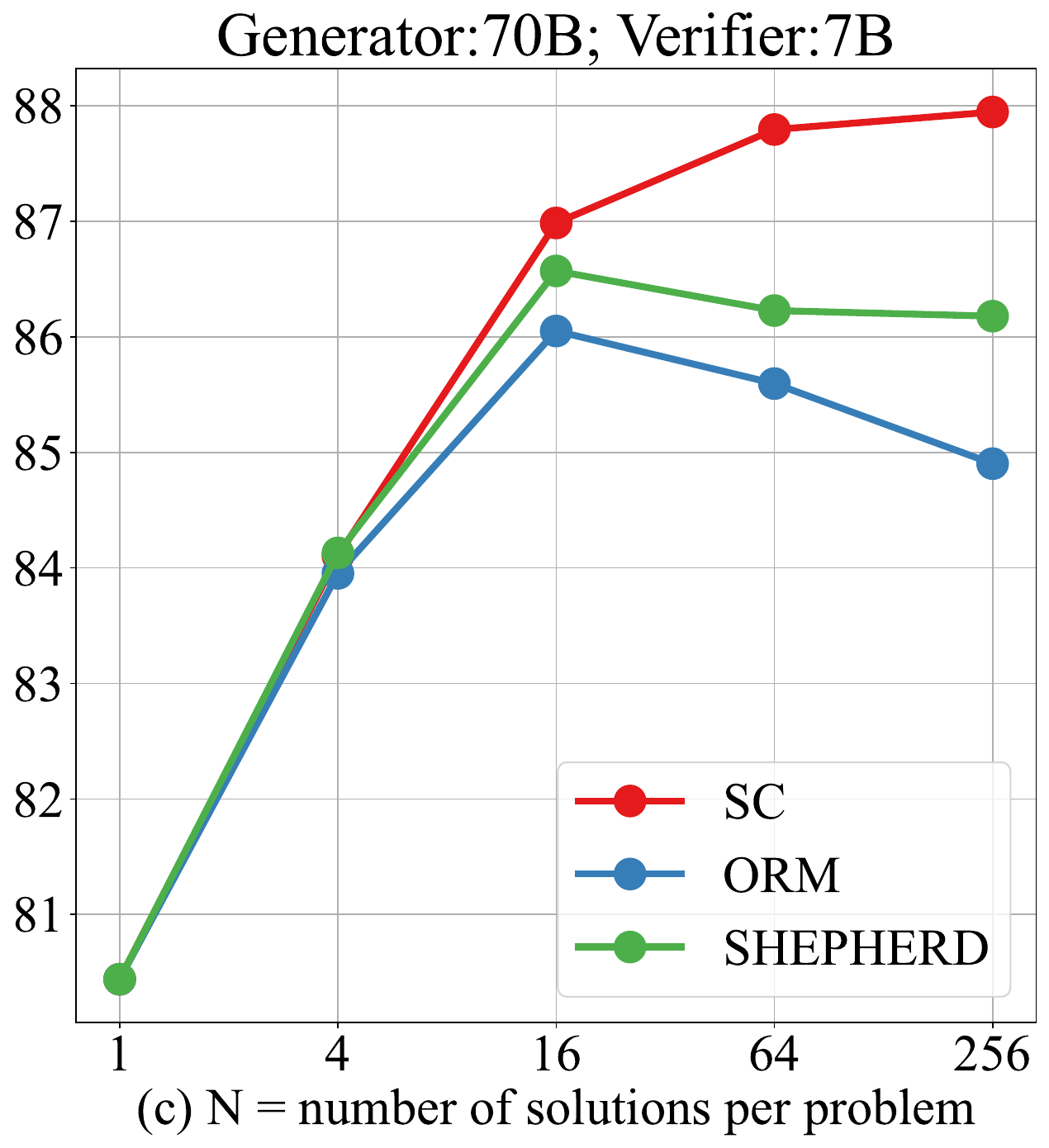}\hspace{-2mm}
}
\subfigure{
\includegraphics[width=0.24\textwidth]{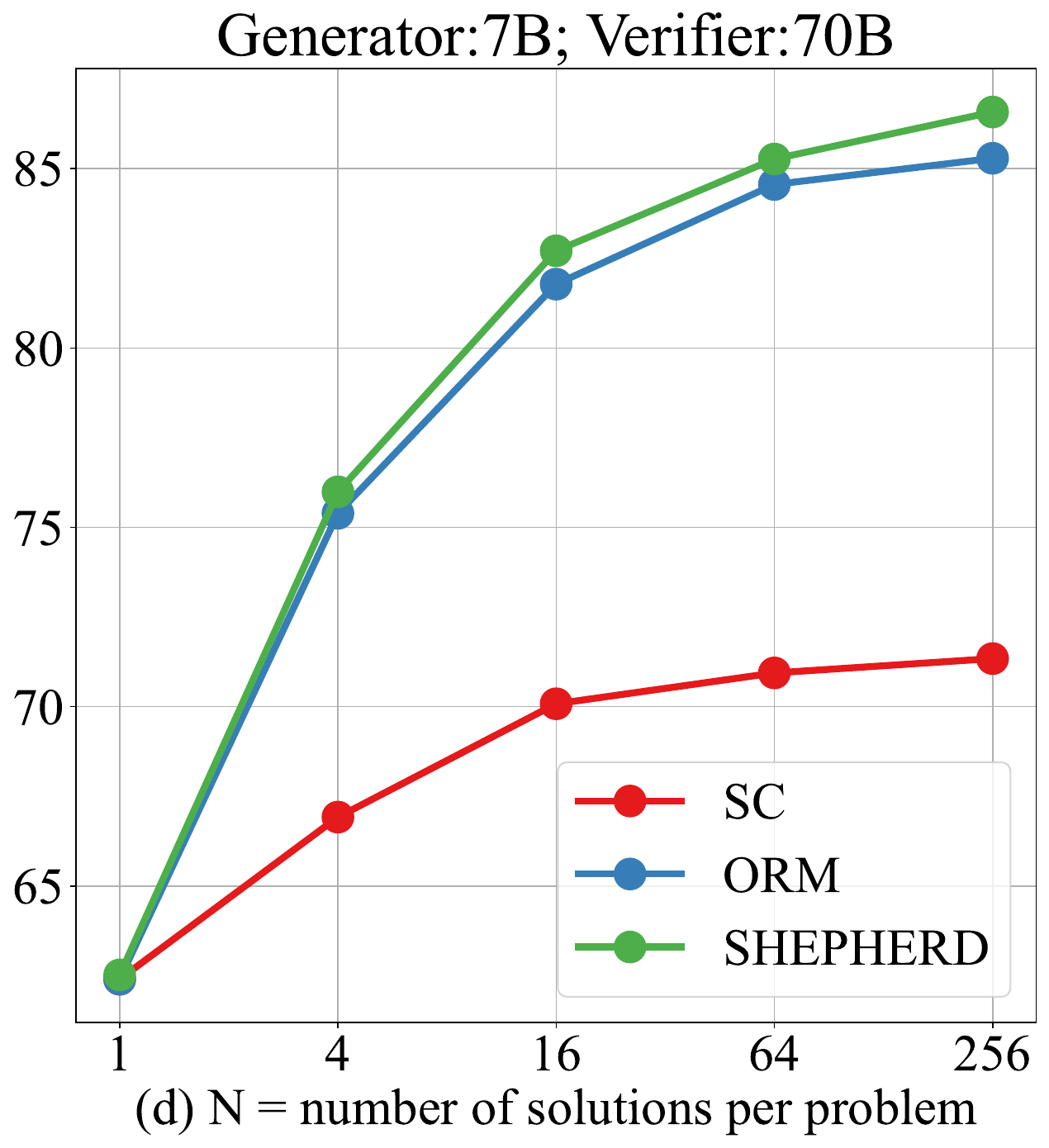}\hspace{-2mm}
}
\caption{Performance of different verification strategies on different sizes of generators and verifiers.}
\label{fig:same_size_model}
\end{figure}

To conduct an exhaustive evaluation of \methodname's effectiveness, we performed a diverse range of experiments using model sizes 7B, 13B, and 70B.

Figures \ref{fig:same_size_model}(a), \ref{fig:same_size_model}(b), and \ref{fig:number-candidate}(a) display the results from the 7B, 13B, and 70B generators paired with equal-sized reward models, respectively. It becomes evident that PRM exhibits superiority over self-consistency and ORM across all sizes of base models. Moreover, bigger reward models prove to be more robust; for instance, the accuracy of the 70B reward models escalates as the number of candidate solutions rises, while the 7B reward models show a decreasing trend.

Figure \ref{fig:same_size_model}(c) and \ref{fig:same_size_model}(d) presents the performance of 7B and 70B generators interfaced with different-sized reward models. The findings illustrate that utilizing a larger reward model to validate the output of a smaller generator significantly enhances performance. Conversely, when a smaller reward model is employed to validate the output of a larger generator, the verification process adversely impacts the model's performance compared to SC.
These results substantiate that we should utilize a more potent reward model for validating or supervising the generator.

\subsection{Influence of the Number of Data} 
\begin{figure}[t]
\centering
\subfigure{
\includegraphics[width=0.48\textwidth]{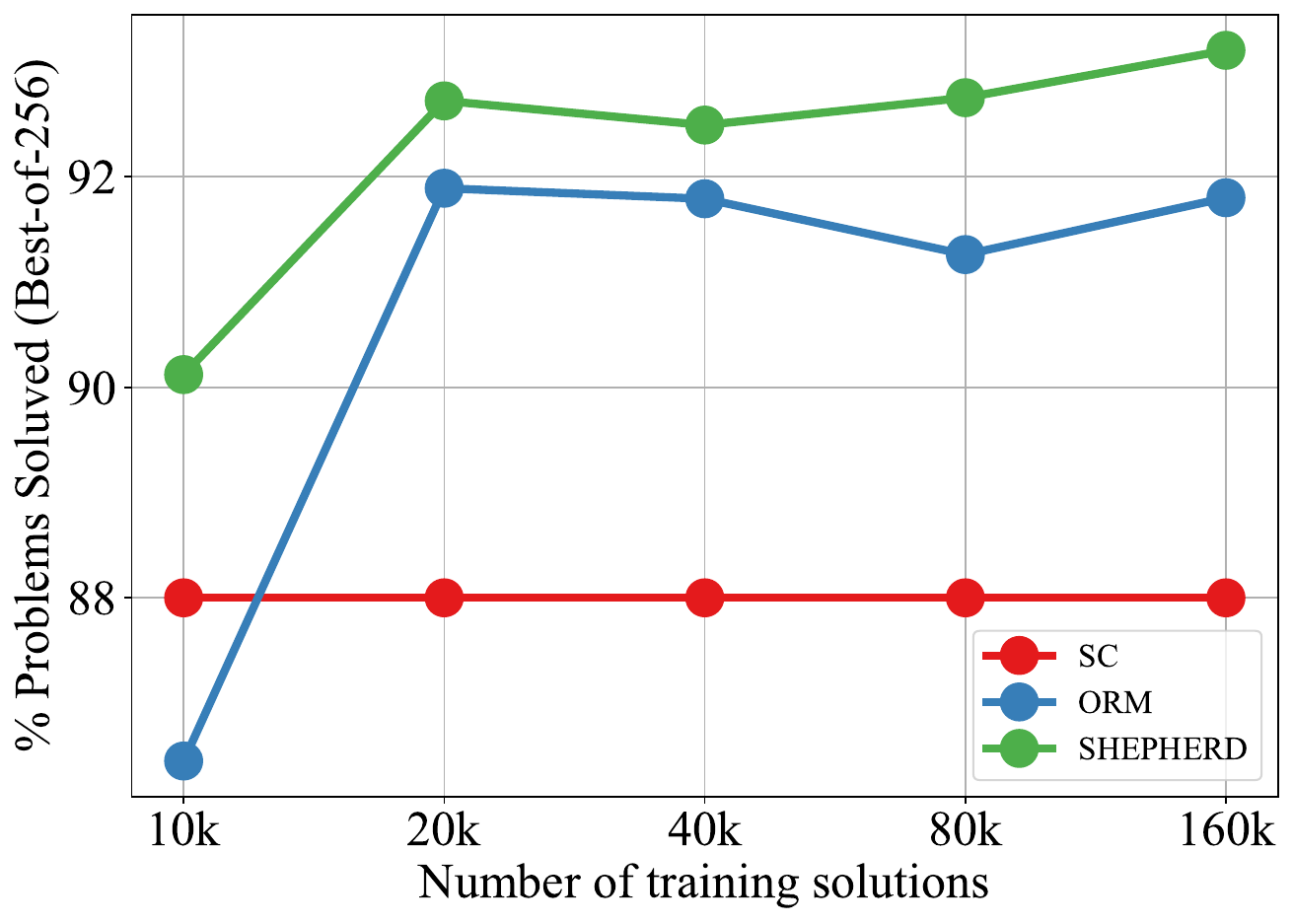}
}
\subfigure{
\includegraphics[width=0.48\textwidth]{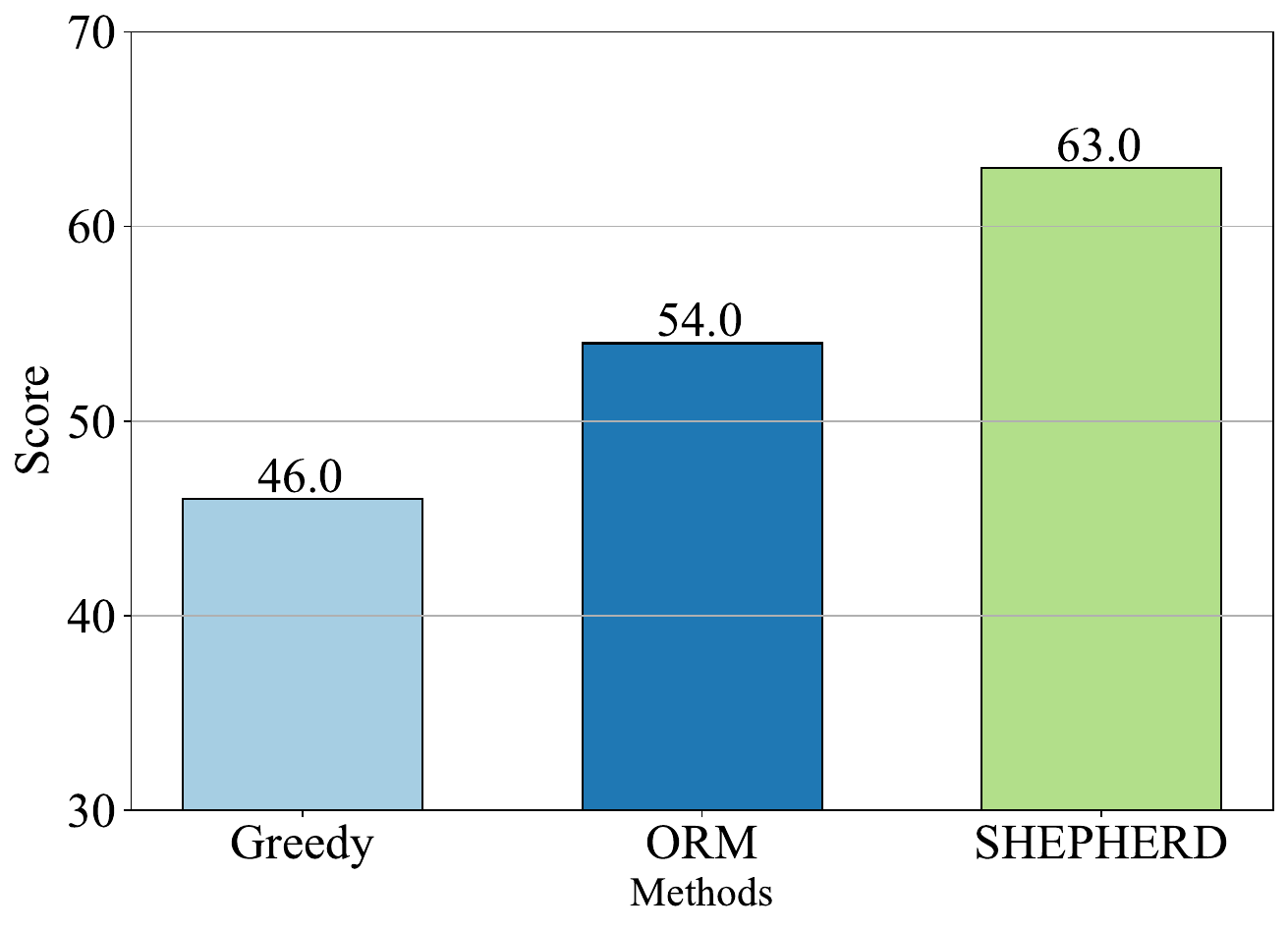}
}
\caption{(a): Performance of different reward models using different numbers of training data; (b) performance of different verification strategies on the out-of-distribution Hungarian national exam.}
\label{fig:number_and_ood}
\end{figure}

We delve deeper into the analysis of PRM and ORM by utilizing varying quantities of training data. As depicted in Figure \ref{fig:number_and_ood}(a), it is clear that PRM exhibits superior data efficiency. Specifically, it outperforms ORM by approximately 4\% accuracy when applying a modestly sized training dataset (i.e., 10k instances). Furthermore, PRM seems to have a higher potential ceiling than ORM. These observations highlight the efficacy of PRM for verification purposes.

\subsection{Out-of-distribution Performance}
To further demonstrate the effectiveness of our method, we conduct an out-of-distribution evaluation on the Hungarian national final exam\footnote{\url{https://huggingface.co/datasets/keirp/hungarian_national_hs_finals_exam}}, which consists of 33 questions. The total score of these questions is 100.
We use the LLemma-34B trained on MetaMATH to serve as the generator and generate 256 candidate solutions for each question. We use LLemma-34B-ORM and LLemma-34B-PRM to select the solution for each question.
As shown in Figure \ref{fig:number_and_ood}(b): 1) both  LLemma-34B-ORM and LLemma-34B-PRM outperform the origin LLemma-34B, showing the reward model can generalize to other domains; 2) PRM outperforms ORM 9 scores, further demonstrating the superiority of PRM. 

We also conduct a case study to intuitively demonstrate the effectiveness of \methodname. 
As outlined in Table \ref{tab:case}, when presented with a question from the Hungarian national final exam, our \methodname accurately selected the correct solution from a pool of 256 potential solutions, which ORM failed.
Moreover, \methodname displayed superior discernment by precisely identifying incorrect steps within the solutions selected by ORM. Notably, it recognized errors in Step 2, Step 6, and Step 9 and so on, and subsequently assigned them lower scores relative to those for steps present in the correct solutions.

\begin{table*}[!tp]
\setlength\tabcolsep{4pt}
\scriptsize

\centering
    \begin{tabular} {p{1.5cm}|p{11.7cm}}
    % \begin{tabular}{l|l|l}
        \toprule
         \textbf{Question} & \textit{The first term of an arithmetic sequence is 18. The sum of the first six terms of the sequence is equal to the sum of the first seven terms. Show that the sum of the first 13 terms of the sequence is 0 and calculate the 13th term of the sequence.}
         \\
         \midrule
         \multirow{3}{*}{\rotatebox{40}{\methodname}} & Step 1: Let the common difference of the arithmetic sequence be $d$.  \textcolor{teal}{\methodname: 0.99}. \\
& Step 2: The sum of the first six terms of the sequence is $18 + (18+d) + (18+2d) + (18+3d) + (18+4d) + (18+5d) = 108 + 30d$. \textcolor{teal}{\methodname: 0.85}.\\
& ...... \\
 & Step 6: Therefore, $d = -3$. \textcolor{teal}{\methodname: 0.85}. \\
& ...... \\
& Step 8: The 13th term of the sequence is $18 - 36 = -18$. \textcolor{teal}{\methodname: 0.93}. \textcolor{olive}{ORM Score: 0.73}.\\
         \midrule
         \multirow{6}{*}{\rotatebox{40}{ORM}} & Step 1: Let the common difference of the arithmetic sequence be $d$.  \textcolor{teal}{\methodname: 0.99}. \\
& Step 2: The sum of the first six terms is $18 + (18 + d) + (18 + 2d) + (18 + 3d) + (18 + 4d) + (18 + 5d) =$  \textcolor{red}{90} + $15d$. \textcolor{teal}{\methodname: 0.53}.\\
& ...... \\
& Step 6: Dividing by $-6$, we find that $d = -2$. \textcolor{teal}{\methodname: 0.38}.\\
& ...... \\
& Step 9: The 13th term of the sequence is $18 - 26 = -8$. \textcolor{teal}{\methodname: 0.38}. \textcolor{olive}{ORM Score: 0.84}.\\
         \midrule

    \end{tabular}
    \caption{A case study from the Hungarian national exam. \textcolor{red}{Red} text denotes the mistake that ORM fails to detect.}
    \label{tab:case}
\end{table*}
\section{Limitations}
Our paper has some limitations, which we leave for future work:

\paragraph{The computational cost of the completion process.} To determine the label of each reasoning step, we utilize a `completer' to decode N subsequent reasoning processes. We observe that as N increases, so does the quality of automatic annotations. However, this completion process demands a lot of computing resources, potentially imposing a limitation on the usage of our method. Despite this limitation, the cost remains significantly lower than human annotation. 
Furthermore, we are optimistic that advancements in efficient inference techniques such as speculative decoding \citep{xia2022lossless,leviathan2023fast} and vLLM \citep{kwon2023efficient} could mitigate this limitation.

\paragraph{The automatic process annotation consists of noise.} Similar to the automatic outcome annotation,  our automatic process annotation also has noise.
Despite this, our experiments verify the efficacy of our method for training a PRM. In particular, the PRM trained on our dataset outperforms the human-annotated PRM800K dataset. 
However, a noticeable gap remains between PRM800K and the candidate responses generated by the open-source models utilized in this study, which may result in the invalidation of PRM800K. 
As a result, the impact of this potential noise on PRM performance is still undetermined. A comprehensive comparison between human and automated annotations is envisaged for future studies. Furthermore, we assert that integrating human and automated process annotations could play a vital role in constructing robust and efficient process supervision.

\section{Conclusion}
 In this paper, we introduce a process-oriented math verifier called \methodname, which assigns a reward score to each step of the LLM's outputs on math problems. The training of \methodname is achieved using automatically constructed process-wise supervision data, thereby eradicating the necessity for labor-intensive human annotation. Remarkably, this automatic methodology correlates strongly with human annotations. Extensive experiments in both verification and reinforcement learning scenarios demonstrate the effectiveness of our method.

\bibliography{PRM}
\bibliographystyle{iclr2024_conference}

\end{document}